\documentclass[letterpaper, 10 pt, conference]{ieeeconf} 
\IEEEoverridecommandlockouts     
\usepackage{cite}
\usepackage{amsmath,amssymb,amsfonts}
\usepackage{algorithmic}
\usepackage{graphicx}
\usepackage{textcomp}
\usepackage{xcolor}
\usepackage{algorithm}
\usepackage{hyperref}
\usepackage{cleveref}

\def\BibTeX{{\rm B\kern-.05em{\sc i\kern-.025em b}\kern-.08em
    T\kern-.1667em\lower.7ex\hbox{E}\kern-.125emX}}
\begin{document}

\title{\LARGE \bf Obstacle Avoidance onboard MAVs \\using a FMCW RADAR}

\author{Nikhil Wessendorp, Raoul Dinaux, Julien Dupeyroux and Guido C.~H.~E. de Croon\authorrefmark{1}

\thanks{\authorrefmark{1}All authors are with Faculty of Aerospace Engineering,  Delft  University  of  Technology,
        Kluyverweg 1, 2629HS Delft, The Netherlands.
        {\tt\small j.j.g.dupeyroux@tudelft.nl}}
}

\maketitle
\thispagestyle{empty}
\pagestyle{empty}

\begin{abstract}
Micro Air Vehicles (MAVs) are increasingly being used for complex or hazardous tasks in enclosed and cluttered environments such as surveillance or search and rescue. With this comes the necessity for sensors that can operate in poor visibility conditions to facilitate with navigation and avoidance of objects or people. Radar sensors in particular can provide more robust sensing of the environment when traditional sensors such as cameras fail in the presence of dust, fog or smoke. While extensively used in autonomous driving, miniature FMCW radars on MAVs have been relatively unexplored. This study aims to investigate to what extent this sensor is of use in these environments by employing traditional signal processing such as multi-target tracking and velocity obstacles. The viability of the solution is evaluated with an implementation on board a MAV by running trial tests in an indoor environment containing obstacles and by comparison with a human pilot, demonstrating the potential for the sensor to provide a more robust sense and avoid function in fully autonomous MAVs.
\end{abstract}

\section{Introduction}
\label{sec:introduction}
\noindent Micro Air Vehicles (MAVs) are very well suited for navigation in complex environments such as indoor buildings as a result of their lightweight, compact design and manoeuvrability, making them ideal for tasks such as search and rescue in hazardous environments and surveillance. To ensure safe flight in such environments, the MAV is usually required to reach a destination while also sensing and avoiding (S\&A) obstacles or people. Apart from dealing with cluttered, GPS-denied environmental conditions, closed tight spaces and limited visibility, MAVs are also constrained by computational, power and weight limitations. For these reasons, the use of cheap, lightweight, and passive vision systems are among the most popular methods. Although cameras are a rich source of information, they also demand an adequate amount of computational power and sufficient visibility conditions. In the absence of these requirements, for example when providing aid and assistance in a smoke-filled building, other sensors need to be considered for a more robust solution to guarantee operation and safety.

In low-light conditions, event-based cameras, laser-based sensors and illumination can compensate for the deficit left by ordinary cameras. However, these systems quickly break down in the presence of dust, fog or smoke. To combat this, ultrasound (sonar) sensors or radar sensors can be utilised instead. Ultrasound sensors are however point-based and have difficulty sensing soft or curved edges at large incidence angles \cite{ultra_Borenstein1988}. Radar sensors on the other hand have been used extensively in the last century for object tracking in the aerospace industry. However these sensors have traditionally been expensive, complex, heavy and power hungry. Only recently have all these factors been improved upon to produce cheap, lightweight sensors that are typically used as auxiliary sensors for applications such as advanced driver assistance systems and ground based applications. These new sensors come in the form of compact millimetre-wave (MMW) frequency-modulated continuous-wave (FMCW) radars. These radars provide the range, bearing and radial velocity of detections \cite{FMCW2}, which can be used for the purpose of multi-target tracking (MTT) and ultimately avoidance. Usage of the radar sensor on small MAVs for indoor obstacle avoidance has not properly been established yet, and the rare existing work does not allow for proper bench-marking.

This study will attempt to fill this gap. While physically ideal for use on small MAVs, the major drawback lies with the fact that the sensor is noisy, and thus requires fine-tuning and filtering to extract meaningful data and perform S\&A functions reliably. The challenge that this paper addresses is to implement such a filtering and tracking pipeline appropriate for real-time use on a MAV. We evaluate the avoidance capabilities in a flight arena equipped with the OptiTrack motion tracking system, and test is on a dataset\footnote{\url{https://github.com/tudelft/ODA_Dataset}} acquired with ground truth positions in the same arena, containing obstacle avoidance trials.

\Cref{sec:related} brings into view the existing work done on FMCW radars and integration onto Unmanned Aerial Vehicles (UAVs). \Cref{sec:std} will explain the processing pipeline and avoidance algorithms used, followed by \Cref{sec:benchmark} giving the results and performance of the implementation.

\section{Related Work}
\label{sec:related}
\noindent The underlying technology and processing pipeline for radars are well explored. While radar sensors are preferably not used independently due to their lack of fidelity, a lot of research has been done to fuse target information with vision systems. Long et al. 2019 \cite{8Long2019} for instance develop a system to aid the visually impaired, which utilises a particle filter to both fuse information and track objects using FMCW radar, a normal camera (using a convolutional neural net) and a stereoscopic IR camera setup, combining the advantages of vision (object classification and identification) with the accurate range, bearing and radial velocity measurements provided by the FMCW radar. Kim et al. 2014 \cite{10Kim2014} and Ćesić et al. 2016 \cite{11cesic2016} also fuse vision and FMCW radar for the purpose of MTT by means of an association algorithm (joint probabilistic data association filter - JPDAF) in combination with a Kalman filter (KF). Both methods perform similar functions, however a particle filter takes a fully probabilistic Bayesian approach, while still demonstrating computational tractability and convergence \cite{9Kreucher2003}.

With regards to airborne applications, little has been explored concerning close proximity obstacle avoidance, although some studies address the use of a (often heavier and more complex) radar as a complimentary sensor to the Traffic Collision and Avoidance System (TCAS) and Flight Alarm (FLARM) for integrating UAVs into the local airspace \cite{1Allistair2011,2Kwag2007,3Kemkemian2009}. Eric et al. 2013 \cite{14Eric2013} achieve this by only utilising a stand-alone FMCW radar sensor using beamforming for an increased field of view (FOV), however outdoor domain of this application is quite different in nature from an indoor environment. Scannapieco et al. 2015\cite{15Scannapieco2015} instead use a gimbaled 94GHz FMCW radar for mapping an indoor environment using Interferometric Synthetic Aperture Radar, and subsequently using that for path planning, however requiring significantly more time and processing power. Scannapieco et al. 2015 \cite{12Scannapieco2016} evaluate the FMCW radar sensor itself rather than implementation for obstacle avoidance of MAVs, only performing outdoor ground tests, and Yu et al. 2020 \cite{6Yu2020} fuse information from a camera and FMCW radar for obstacle avoidance, however are still very reliant on the vision system.

\section{Signal Processing}
\label{sec:std}
\subsection{Detection}
\begin{figure}
    \centering
    \includegraphics[width=0.75\linewidth]{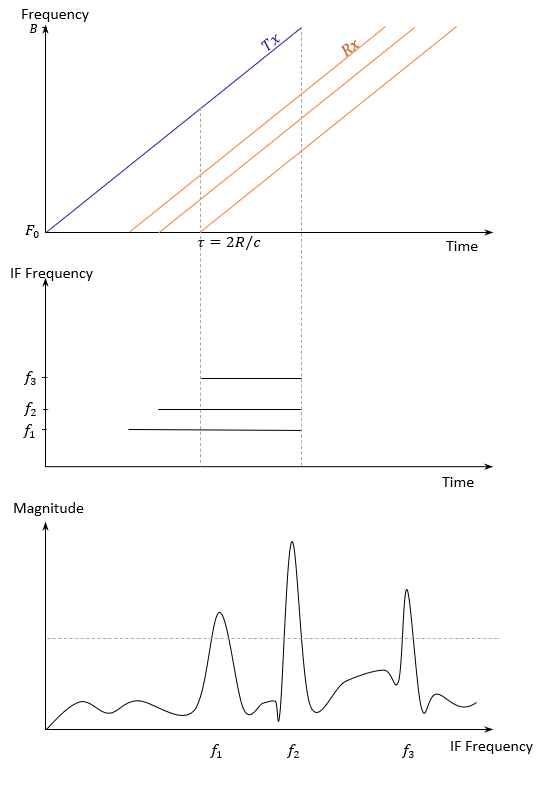}
    \caption{The basic principle behind FMCW radar. Top: the transmitted (purple) and received (orange) waves of one chirp. Middle: the mixed IF signal, showing the distinct frequencies that different objects produce. Bottom: the first range FFT (only the magnitude shown) applied to the IF signal, delineating the peaks.}
    \label{fig:FMCW}
\end{figure}

\noindent Specifically, the sensor in use is a fast-chirp FMCW radar and operates by transmitting a saw-tooth FM carrier wave from its transmitter antenna and listening for the returns reflected by objects in its two receiver antennas (one period is referred to as a 'chirp'). The received signals (one per object) are shifted to the right (a delay in time) with increasing range, as shown in \Cref{fig:FMCW}. The transmitted and reflected signals are then mixed to produce the intermediate frequency (IF) signal (or beat frequency) and is passed through a low pass filter followed by an ADC to produce the raw data of the radar, which consists of the $(I,Q)$ values representing the electromagnetic wave. Fast-chirp FMCW radars feature reduced range but better resolution compared with the traditional FMCW radars, where the chirp duration is one order magnitude longer. Note that because of this fast-chirp nature, the Doppler shift in frequency is negligible compared to the shift in frequency due to range, and is not accounted for in this step.

Following from this raw data, the IF signal is passed through a fast-Fourier transform (FFT) (with zero padding), which highlights the peaks representing range produced by all objects in the FOV. Once a threshold is applied, the detections are distinguished with their associated range using \Cref{eq:range}:

\begin{equation}
    R = \frac{c T_c f_b }{2B} = \frac{c f_b }{2S}
    \label{eq:range}
\end{equation}

\noindent where c is the speed of light, B is the bandwidth, $f_b$ is the beat frequency, S is the slope of the frequency modulated ramp and $T_c$ is the up-chirp time. To determine the horizontal bearing of the detections, the phases of the two antennas in the FFT (where only magnitude is shown in \Cref{fig:FMCW}) are compared with the use of \Cref{eq:bearing}:

\begin{equation}
\theta= \arcsin \left( \frac{\lambda \Delta \omega_d}{2\pi d} \right)=\arcsin \left( \frac{ \Delta \omega_d}{\pi} \right)
\label{eq:bearing}
\end{equation}

\noindent where d is the antenna spacing, $\lambda$ is the wavelength and $\Delta \omega_d$ is the phase difference between the two antennas. Note that $d=0.5\lambda$ gives the largest FOV of $\pm90^{\circ}$. 

The radial velocity of the detected objects can also be extracted by taking a second set of FFTs over multiple chirps: this essentially compares the change of phase over 2 consecutive chirps, since the phase is very sensitive to small changes in distance (essentially, a phase change is a Doppler frequency shift). This change in phase $\Delta \omega$ is given by \Cref{eq:phase}, where V is the radial velocity of the target and $\lambda$ is the wavelength.

\begin{equation}
    \Delta \omega = \frac{4 \pi V T_c}{\lambda}
    \label{eq:phase}
\end{equation}

\begin{figure}[]
    \centering
    \includegraphics[width=1\linewidth]{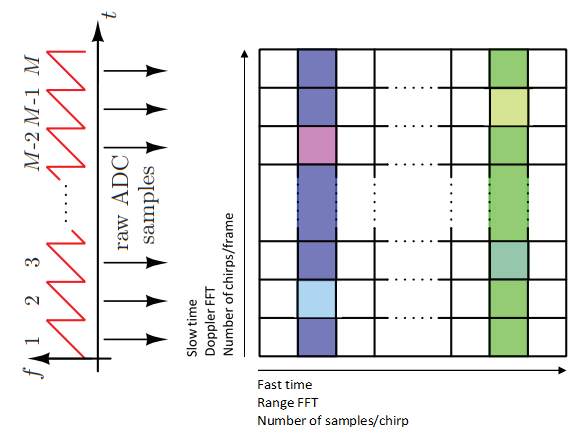}
    \caption{A 2D FFT, where the horizontal axis represents fast time (one chirp) and the vertical axis represents slow time (one frame). After performing FFTs on the fast time axis identifying any peaks (shaded regions), an FFT is performed along the slow time axis to determine the phase change (Doppler shift) of same range bin can be compared to distinguish one or multiple peaks (objects with different velocities at same range, coloured squares). Adapted and modified from \cite{FMCW2}.}
    \label{fig:2dfft}
\end{figure}

By taking a number $M$ of IF samples instead of 2 (number of chirps in a frame), a velocity estimate can be computed for each individual object by taking a Doppler FFT of $\Delta \omega$ over the different chirps, creating a 2D FFT matrix shown in \Cref{fig:2dfft}. Here objects can be resolved by both their range and radial velocity.

\subsection{Filtering}
\noindent As the sensor is rather noisy, both data association and tracking have to be employed. Data association involves handling the detections and objects that are being tracked, that is, first: assigning detections to existing objects and discarding detections from clutter, second: creating new objects when detections indicate there is a new object in the FOV, and third: deleting objects when they leave the FOV (when there are no new detections). These operations are done in conjunction with the Kalman filtering (KF) process.

The first process boils down to calculating a cost matrix which indicates the cost of associating a detection to an object or clutter. For this, a simple global nearest neighbourhood (GNN) optimisation algorithm is used \cite{GNN_Cox1993}, which defines the cost as being proportional to the square of the distance between the detection and prediction of the object position (using the KF). The cost matrix also contains the cost associated with misdetecting an object, that is, no detections associated with the object (right side of \Cref{eq:costmatrix} which shows the cost matrix $L$). Additionally, gating is used, whereby any detections that are greater than a threshold distance to a particular object are immediately discarded ($-\ell^{n, m}$ = $\infty$). The GNN algorithm is a greedy yet computationally efficient approach that works well for simple scenarios. This cost matrix is then converted to an assignment matrix by minimising the cost using an algorithm such as the Hungarian algorithm \cite{sahbani2016kalman}.

\begin{equation}
    L=\left[\begin{array}{ccc|ccc}
-\ell^{1,1}  & \ldots & -\ell^{1, m} & -\ell^{1,0}  & \ldots & \infty \\
\vdots & \ddots & \vdots & \vdots & \ddots & \vdots \\
-\ell^{n, 1} & \ldots & -\ell^{n, m} & \infty & \ldots & -\ell^{n, 0}
\end{array}\right]
\label{eq:costmatrix}
\end{equation}

\begin{equation}
    \ell^{i, 0, h}=\log \left(1-P^D\right)
\end{equation}
\begin{equation}
    \ell^{i, j, h}=-\frac{1}{2}\left(z^{i}-\hat{z}^{i, h}\right)^{\top}\left(S^{i, h}\right)^{-1}\left(z^{j}-\hat{z}^{i, h}\right)
    \label{eq:costelem}
\end{equation}

\noindent where $n$ is the number of objects being tracked, $m$ is the number of radar detections, $-\ell^{n, m}$ represents the association cost, and $-\ell^{n, 0}$ represents the cost of misdetecting the object. $P_d$ is the probability of detection, $z^{i}-\hat{z}^{i, h}$ is the distance between the measurement and predicted location of the object, and $S^{i, h}$ is the innovation covariance of the KF. The second step is done by keeping track of all detections within the FOV (associating them to new candidate tracks which are also tracked with a KF). When the covariance of the position (in the $P^{i, h}$ matrix of the KF) drops below a threshold, the object is initiated (track birth) and considered valid. Likewise, when a tracked object's covariance rises above another threshold (when it is misdetected multiple times) it is removed (track death).

Once a detection has been associated with an object, the detection is used as the measurement input (range, bearing, radial velocity) to an ordinary KF that is run for every object to filter out noise and estimate the tangential velocity as well, thereby obtaining the range, bearing and their derivatives. The KF assumes both observation noise, to account for the sensor noise, and process noise, to account for any non-linearities in the motion of the objects or MAV, as a constant acceleration model is assumed.

\subsection{Avoidance}
\noindent The obstacle avoidance control method used is Velocity Obstacles (VO), which finds the set of velocity vectors of the MAV that will result in a collision with the object, taking into account the radius of both the object and MAV. 

\begin{figure}
  \centering
  \includegraphics[width=0.8\linewidth]{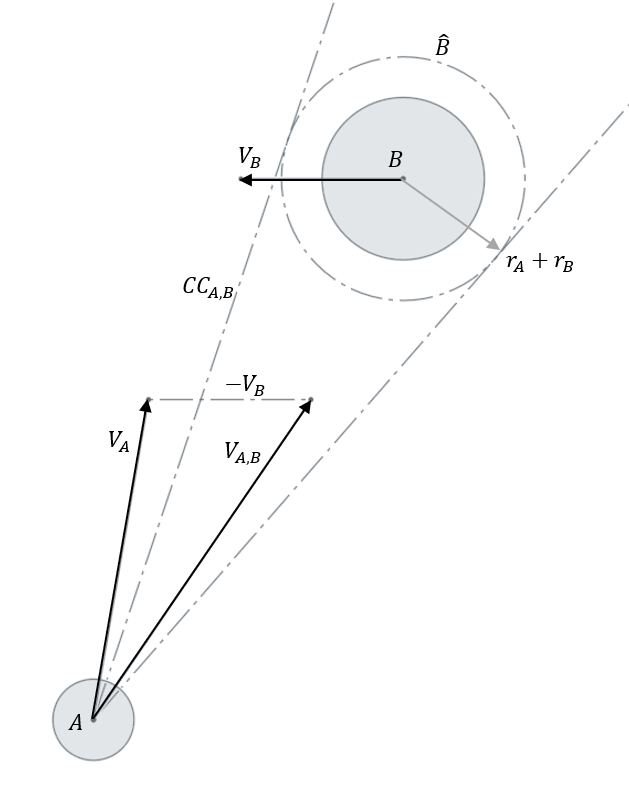}  
  \caption{Robot $A$ and moving obstacle $B$ will collide as $V_{A,B}$ lies within $CC_{A,B}$, which is formed by enlarging object with the radius of robot.}
  \label{fig:VOa}
\end{figure}

The following explanation is retrieved from Fiorini et al. 1998 \cite{FioriniVO2}. Consider a robot $A$ and an obstacle $B$ with velocities $V_A$ and $V_B$ and radii $r_A$ and $r_B$, as shown in \Cref{fig:VOa}. Mapping B onto the configuration space of A means enlarging object $B$ by the radius of $A$ to form object $\hat{B}$, and reducing A to a point $\hat{A}$ and computing the relative velocity of robot $A$ with respect to object $B$, $V_{A,B} = V_A - V_B$. The collision cone $CC_{A,B}$ can then be formed, in which any \textit{relative} velocity $V_{A,B}$ will result in a collision with object $B$. The radar sensor will yield the relative position and $V_{A,B}$. Since the ego-velocity $V_{A}$ is known, $V_{B}$ can also be determined.

By accounting for the velocity of robot A and its limitations in maximum velocity and direction change, a desired $\hat{V}_{A,B}$ can be computed by adjusting $V_{A,B}$ to lie on one of the edges of $CC_{A,B}$ (also taking into account any safety margins). An absolute desired velocity of robot A, $\hat{V}_{A}$, can then be found by addition with $V_{B}$. In the case in \Cref{fig:VOa}, it is most beneficial to slow down and adjust the velocity vector to the right.

\subsection{Sensor Characteristics}

\noindent The sensor used in this study is the Infineon XENSIV$^{TM}$ 24GHz Position2Go kit, a small 10g fast-chirp FMCW radar that features human target detection at a range of 1-12m and a horizontal-vertical half-power beamwidth (HPBW) FOV of $76^{\circ} $x$ 19^{\circ}$. Although this is sufficient for frontal obstacle avoidance, objects that are moving faster than the MAV outside the FOV still pose a collision threat, albeit less likely. With a maximum bandwidth of 200MHz, it is able to resolve objects $0.75m$ apart in range, with a range accuracy of $\pm$ 15cm and an angular accuracy of $\pm 2^{\circ}$ from $0-20^{\circ}$, and up to $\pm 8^{\circ}$ from $20-65^{\circ}$. Strict filtering of clutter detections is required due to the noisy nature of the sensor. Furthermore, as the sensor only features 2 receiver antennas, and is thus only able to detect 2 objects at a time in the same range bin. However taking this into account in the association algorithm, all objects in the FOV can be detected over multiple frames (although decreasing the update frequency).

\section{Performance Evaluation}
\label{sec:benchmark}
\subsection{Implementation}
\begin{figure}
    \centering
    \includegraphics[width=\linewidth]{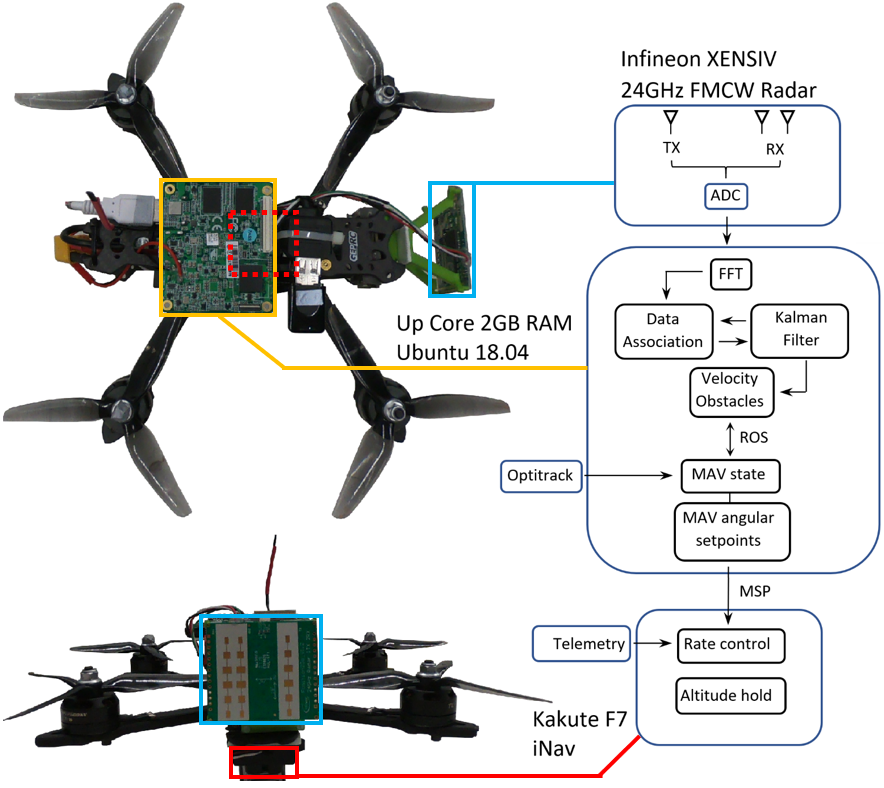}
    \caption{Top and front view of the MAV. Light blue indicates the FMCW radar, yellow the Up Core companion computer, and red the flight controller (underneath the companion computer) and the LiDAR altimeter.}
    \label{fig:hardware}
\end{figure}

\noindent The FMCW radar sensor was integrated and tested on a custom made 5-inch MAV, as shown in \Cref{fig:hardware}. Two processing boards are integrated. The first is the Kakute F7 flight controller running iNav 2.6.0 firmware, the second companion computer is the Intel Up Core (1.44GHz 64bit processor with 2GB RAM) running Ubuntu 18.04 LTS. The latter runs the radar driver, processes the raw data, performs the MTT and runs a custom made autopilot using ROS (Robot  Operating  System) to communicate with the radar sensor, computing the avoidance manoeuvre and relaying the desired orientation of the MAV (pitch, roll and yaw angles) to the flight controller (using MSP protocol), which in turn handles the low level rate and altitude control using the TFMini LiDAR rangefinder facing down. The radar sensor is fixed to the front of the MAV at a slight upward tilt of $10^{\circ}$ to reduce reflections from the ground.

\noindent Testing was done in the flying arena of the TU Delft, equipped with the OptiTrack motion capture system for positioning, which is relayed through UDP to the UP Core, although concerning the avoidance algorithm, only velocity control was implemented. Furthermore, the avoidance algorithm only considers the nearest obstacle, both for simplicity and to stimulate different avoidance scenarios. A simpler avoidance manoeuvre was implemented whereby the MAV simply translates approximately 1m to the side to better approximate the flying behaviour displayed in the dataset in which a human is flying to avoid 1 or 2 obstacles in the flying arena. The obstacles are cardboard poles roughly 0.5m in diameter placed in the centre of the flying arena, and avoidance was carried out from all sides and corners.

\subsection{Results}
\begin{figure}[!t]
    \centering
    \includegraphics[width=0.9\linewidth]{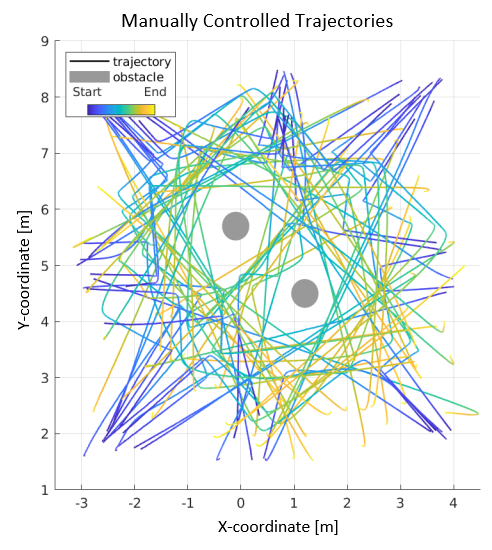}
    \caption{Manually flown trajectories of the MAV of 78 samples from the obstacle avoidance dataset, avoiding 2 obstacles from different angles.}
    \label{fig:traj_man}
\end{figure}

\begin{figure}[!t]
    \centering
    \includegraphics[width=0.9\linewidth]{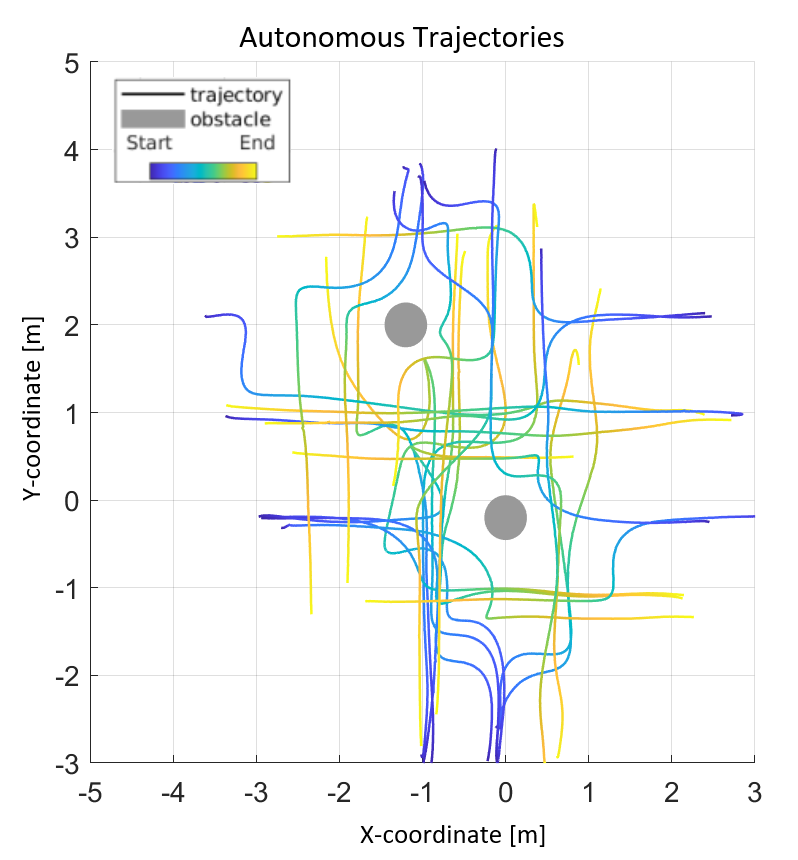}
    \caption{26 sample trajectories of the autonomously controlled MAV using the radar sensor.}
    \label{fig:Traj_auto}
\end{figure}

\begin{figure}[!t]
    \centering
    \includegraphics[width=0.9\linewidth]{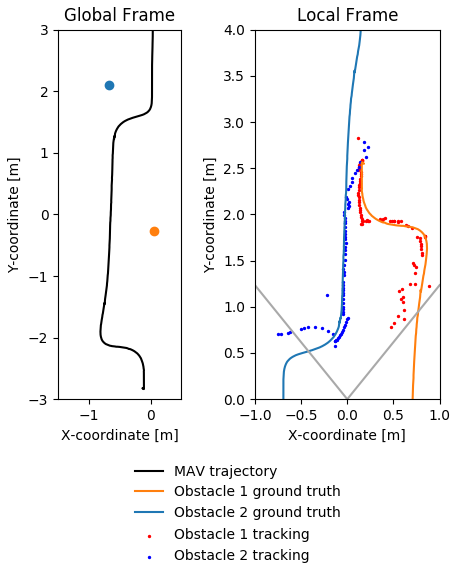}
    \caption{Illustration of what the FMCW radar sensor detects after filtering and tracking (right) when following the sample trajectory on the left. The grey lines indicate the HPBW FOV of the radar ($78^{\circ}$) and the ground truth position is shown in both figures (orange and light blue paths)}
    \label{fig:radar_path}
\end{figure}

Looking at \Cref{fig:traj_man} and \Cref{fig:Traj_auto}, showing the trajectories taken when the MAV is controlled by a human pilot versus the on board obstacle avoidance controller using the radar, it is evident that the FMCW radar can reliably detect obstacles and determine when a collision is imminent, thus allowing the MAV to safely avoid damage or injury to the MAV or environment. On occasion the radar will struggle to track the further obstacle due to the inherent noise of the sensor, however when brought close enough to the obstacle the MAV was still able to perform a successful avoidance manoeuvre. This can best be visualised in \Cref{fig:radar_path}, which on the left shows the ground truth trajectory and location of the MAV and obstacles, and on the right the output of the filtering and tracking algorithms, showing the relative paths taken by the obstacles (Doppler information is not displayed). First obstacle 1 comes into view (orange ground truth and red detections), which the MAV avoids by moving to the left (or the obstacle moving to the right relative to the MAV), followed by the second obstacle coming into view approximately 0.7 seconds later (light blue ground truth and dark blue detections), which the MAV avoids to the right.

\begin{figure}[!t]
    \centering
    \includegraphics[width=0.9\linewidth]{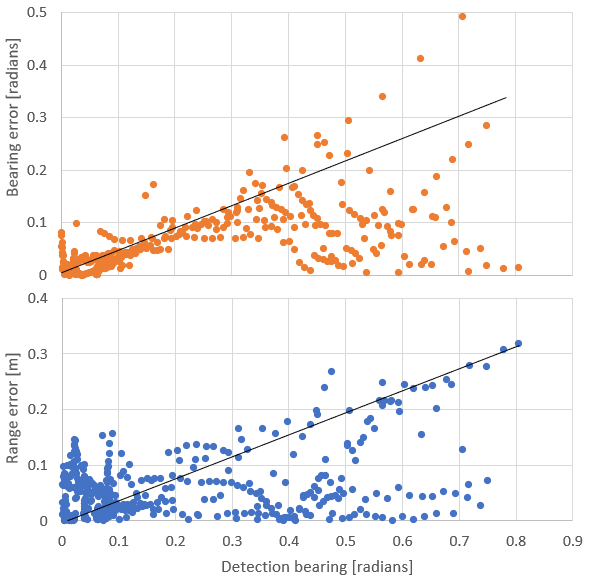}
    \caption{Data from 4 trials showing the approximately linear trend of the error in bearing and range as the object moves further from the centre of the radar.}
    \label{fig:error}
\end{figure}

As can be seen in \Cref{fig:radar_path} on the right, the error in tracking is most evident when the MAV changes trajectory, which in fact represents a non-linearity in the motion of the obstacle (or MAV) meaning it can take some steps before the ordinary KF is able to cope with this. However when the radar sensor would come to a complete halt, an increase in noise around the obstacle was also observed for approximately one second, further exacerbating the error. This however did not impact the MAVs ability to sense and avoid a collision. Additionally, the error in bearing and range increases as the obstacles move towards the edge of the HPBW FOV, which can be seen in \Cref{fig:error}, showing an approximately linear trend.

\section{Conclusion}
\noindent This work demonstrates the pertinence of using a standalone FMCW radar sensor for the purpose of sense and avoid. A multi-target tracking and avoidance algorithm have been implemented on a MAV and tested on both one and two obstacles, showing that the MAV is successfully able to avoid them when solely relying on the radar sensor, demonstrating that reliance on this sensor can be effective when required, especially when other sensors fail due to the presence of fog, smoke or flames. This will ultimately help make MAVs for applications such as surveillance and search and rescue safer and more reliabl

To better detect obstacles in cramped spaces, 77GHz FMCW radars can be used, which feature improved bandwidth and resolution, allowing for more accurate detection of obstacles and perhaps classification of walls as well, however requiring a more robust and computationally expensive data association algorithm capable of clustering detections (e.g. DBSCAN \cite{DBSCAN_Ester1996}). Other FMCW radar sensors also incorporate more than two receiver antennas (allowing for more detections per scan) or beamforming (scanning a larger FOV). While this study has demonstrated that sense and avoid using a standalone radar sensor can be very useful, it is best used when fused with other sensors when circumstances and conditions allow  (even with event-based cameras which can operate in low-light environments, as shown by Zhang et al. 2019 \cite{fusionZhang2019} who fuse the sensors in an EKF to compensate for the error bounds produced by both sensors.

\section*{Supplementary materials}
\label{sec:supplementary}
The ROS implementation our radar-based navigation system can be found here: \url{https://github.com/tudelft/radar_nav}, along with supporting videos. The Obstacle Detection and Avoidance dataset is available at: \url{https://github.com/tudelft/ODA_Dataset}.

\section*{Acknowledgements}
This work is part of the Comp4Drones project and has received funding from the ECSEL Joint Undertaking (JU) under grant agreement No. 826610. The JU receives support from the European Union's Horizon 2020 research and innovation program and Spain, Austria, Belgium, Czech Republic, France, Italy, Latvia, Netherlands.

\bibliographystyle{IEEEtran}
\bibliography{IEEEabrv,IEEEexample}

\end{document}